\documentclass{article}

\PassOptionsToPackage{numbers, compress}{natbib}

\usepackage[preprint]{neurips_2026}
\usepackage[utf8]{inputenc}
\usepackage[T1]{fontenc}
\usepackage{amsmath}
\usepackage{amsfonts}
\usepackage{amssymb}
\usepackage{booktabs}
\usepackage{graphicx}
\usepackage{hyperref}
\usepackage{url}
\usepackage{xcolor}
\usepackage{float}

\title{LiteMedCoT-VL: Parameter-Efficient Adaptation for Medical Visual Question Answering}

\begin{document}

\author{%
    Runze Ma \\
    School of Information Technology\\
    Monash University Malaysia\\
    \texttt{rmaa0033@student.monash.edu} \\
    \And
    Shunbo Jia \\
    Faculty of Innovation Engineering\\
    Macau University of Science and Technology\\
    \texttt{2240003657@student.must.edu.mo} \\
    \And
    Haonan Lyu \\
    Department of Bioelectronics\\
    Faculty of Biomedical Engineering\\
    Shenzhen University of Advanced Technology\\
    \texttt{SUAT25060153@stu.suat-sz.edu.cn} \\
    \And
    Guo Liu \\
    School of Mathematics and Statistics\\
    Huazhong University of Science and Technology\\
    \texttt{u202210024@hust.edu.cn} \\
    \And
    Caizhi Liao\thanks{Corresponding author.} \\
    Department of Bioelectronics\\
    Faculty of Biomedical Engineering\\
    Shenzhen University of Advanced Technology\\
    \texttt{liaocaizhi@suat-sz.edu.cn} \\
}

\maketitle

\begin{abstract}
The reasoning gap between large and compact vision-language models (VLMs) limits the deployment of medical AI on portable clinical devices. Compact VLMs of 2--4B parameters can run on resource-constrained hardware but lack the multi-step reasoning capacity needed for interpretable clinical decision support. Existing knowledge distillation methods transfer answers without the reasoning process behind them. Medical visual question answering (VQA) serves as a testbed for this problem, as it requires models to integrate visual evidence with clinical knowledge through structured reasoning chains. We introduce LiteMedCoT-VL, a pipeline that transfers chain-of-thought reasoning from a 235B teacher model to 2B student models through LoRA-based fine-tuning on explanation-enriched training data. All inference is conducted without image captions by default, simulating the clinical scenario in which a physician interprets a medical image directly without an accompanying radiology report. On the PMC-VQA benchmark, LiteMedCoT-VL achieves 64.9\% accuracy, exceeding the zero-shot Qwen3-VL-4B baseline of 53.9\% by 11.0 percentage points and outperforming all published baselines. This result indicates that a 2B model with reasoning distillation can match or exceed models with twice the parameters. Visual grounding analysis shows that the model relies on image content rather than exploiting textual priors. Our code is publicly available at \url{https://github.com/R4nzer/LiteMedCoT-VL}.
\end{abstract}

\section{Introduction}
\label{sec:introduction}

Medical imaging has progressed from handcrafted features to deep learning and, more recently, to systems that combine visual evidence with language understanding~\cite{zhou2021review,tsuneki2022deep,gao2025medical}. Vision-language models (VLMs) trained on large image--text collections perform well on new tasks without task-specific training~\cite{radford2021learning,jia2021scaling,team2024gemini,zhang2024vision}, and adapting these models for medical image question answering is an active research area~\cite{wang2025large,liu2025visual,sheng2024large}.

Deploying these models on portable medical devices presents a fundamental challenge. Large VLMs achieve strong performance but require computing resources that exceed the capacity of portable imaging equipment and clinical devices with limited hardware. Compact models with 2--4B parameters can run on such hardware but often lack the multi-step reasoning capacity needed for clinical decision support. Knowledge distillation transfers abilities from a large model to a smaller one~\cite{gou2021knowledge,mansourian2025comprehensive}, yet conventional distillation methods transfer only answer labels without the reasoning process behind them.

Recent work has begun to address this gap through structured chain-of-thought (CoT) annotations~\cite{fan2026step,sharma2026chexthought}, reinforcement learning with process rewards~\cite{gulluk2026grpo}, agentic grounding frameworks~\cite{du2026care}, and preference-based optimization~\cite{huang2025elicit}. These approaches demonstrate that transferring reasoning capability, rather than answer patterns alone, improves medical VQA performance. However, they often require expert-annotated reasoning traces, complex multi-stage training pipelines, or substantial additional compute. A simpler recipe that distills reasoning from a large teacher into a compact student via parameter-efficient fine-tuning remains underexplored.

This work introduces LiteMedCoT-VL, a pipeline that adapts compact vision-language models for medical visual question answering through chain-of-thought knowledge distillation and LoRA-based fine-tuning. A large teacher model generates step-by-step clinical reasoning chains, which are injected into the training data to transfer reasoning patterns to a compact student model. The PMC-VQA benchmark~\cite{zhang2023pmc} serves as our evaluation platform.

Our contributions are as follows.
\begin{itemize}
    \item We present LiteMedCoT-VL, a pipeline that transfers chain-of-thought reasoning from a large vision-language model to compact models through parameter-efficient fine-tuning.
    \item We show that chain-of-thought distillation from Qwen3-VL-235B-A22B-Instruct~\cite{bai2025qwen3} improves the accuracy of a 2B parameter student model from 48.7\% to 64.9\% on PMC-VQA.
    \item We report experimental results comparing our approach against published baselines and compact open-source models, demonstrating that our best configuration outperforms all existing methods on this benchmark.
\end{itemize}

\section{Related Work}
\label{sec:related-work}

Medical image analysis has progressed through a sequence of paradigm shifts over the past two decades. Early approaches relied on thresholding, edge detection, and handcrafted feature representations~\cite{jardim2023image,salmanpour2026handcrafted}. The adoption of convolutional neural networks (CNNs) in the 2010s brought substantial improvements in classification and segmentation accuracy~\cite{chen2025review,albuquerque2025deep,mienye2025deep}, and the subsequent introduction of Transformers further advanced global-context modeling for medical imagery~\cite{chen2021transunet,cao2022swin,he2023transformers}. Building on these advances, multimodal approaches that integrate visual evidence with language context emerged as a natural next step, demonstrating strong results across medical imaging tasks~\cite{zhou2021review,gao2025medical}.

This progression gave rise to VLMs that combine visual encoders with language models, enabling image-grounded question answering and report generation~\cite{zhang2024vision,dong2025generative}. Early vision-language pretraining established the foundation through contrastive learning on large-scale image-text collections~\cite{radford2021learning,jia2021scaling}, followed by unified architectures that process both modalities within a single framework~\cite{li2023blip,singh2022flava}. Medical-focused systems adapted these general-purpose models for clinical applications: LLaVA-Med~\cite{li2023llava} and MedICaP~\cite{zhang2023pmc} demonstrated promising results on benchmarks such as SLAKE~\cite{liu2021slake}, PathVQA~\cite{he2020pathvqa}, and VQA-Med~\cite{ben2021overview}. However, the strongest performance consistently came from very large models exceeding 70B parameters, which cannot run on the compute and memory capacity of portable medical devices. This growing gap between model capability and deployability motivated a parallel line of work on model compression and efficient adaptation.

As VLMs grew larger, knowledge distillation emerged as a primary strategy for compressing large models into smaller ones while preserving performance~\cite{gou2021knowledge,mansourian2025comprehensive}. Early methods matched output distributions or intermediate representations between teacher and student~\cite{zheng2024knowledge,yuan2024student}, and have been widely applied in the medical domain to compress models while preserving diagnostic accuracy~\cite{wang2025robust}. In parallel, compact vision-language models were developed with efficient architectures suitable for resource-constrained deployment, including Phi-3.5-vision~\cite{abdin2024phi3}, InternVL2~\cite{chen2024far}, and SmolVLM2~\cite{marafioti2025smolvlm}. Despite these efficiency gains, compact models still lacked the multi-step reasoning capacity of larger systems, creating a need for methods that could transfer reasoning capability rather than just answer patterns.

This need led to combining distillation with chain-of-thought reasoning~\cite{jiang2025comt}, which generates step-by-step explanations that make clinical reasoning explicit. When trained on these explanations, the student model learns not just the correct answer but the reasoning process behind it~\cite{ho2023large,wang2022self}, an approach particularly relevant for medical tasks where interpretable reasoning is as important as the final prediction. Step-CoT~\cite{fan2026step} extends this idea with expert-curated, structured multi-step CoT annotations aligned to clinical diagnostic workflows, coupled with a dynamic graph-structured focusing mechanism. CheXthought~\cite{sharma2026chexthought} provides a large-scale dataset of CoT reasoning traces with synchronized visual attention annotations from hundreds of radiologists, demonstrating that models trained on these traces surpass VLM-generated CoT in factual accuracy and spatial grounding. MedE2~\cite{huang2025elicit} proposes a two-stage pipeline that first elicits multimodal reasoning via orchestrated demonstrations and then enhances it with Direct Preference Optimization. Beyond supervised and preference-based methods, reinforcement learning approaches have emerged: trajectory-aware GRPO~\cite{gulluk2026grpo} applies process rewards based on reasoning-step similarity, and CARE~\cite{du2026care} decomposes medical VQA into specialized sub-modules trained with reinforcement learning with verifiable rewards. These methods demonstrate that reasoning capability can be transferred through diverse mechanisms, though they often require expert annotations, complex multi-stage pipelines, or substantial additional compute.

Full fine-tuning of large foundation models also became impractical in memory and compute, motivating parameter-efficient alternatives. LoRA~\cite{hu2022lora} injects trainable low-rank matrices into attention projections, updating less than 1\% of parameters while preserving pretrained capabilities. Subsequent work introduced adapters~\cite{houlsby2019parameter}, prompt tuning~\cite{li2021prefix}, and hybrid strategies~\cite{mao2022unipelt} to further reduce adaptation cost. In medical settings, these methods enable model customization under the resource constraints of clinical hardware~\cite{mao2025survey,volkov2025visual}. Our approach combines LoRA with chain-of-thought distillation, transferring reasoning capability from a 235B teacher to a 2B student through a simpler, single-stage recipe that does not require expert-annotated reasoning traces.

A growing body of work examines whether medical vision-language models genuinely ground their reasoning in visual evidence. MIRAGE~\cite{asadi2026mirage} demonstrates that models can perform competitively on medical VQA benchmarks even when images are withheld, raising concerns about textual prior exploitation. MedVR~\cite{jiang2026medvr} proposes annotation-free visual reasoning through agentic reinforcement learning with entropy-guided visual regrounding. ViTAR~\cite{chen2025think} introduces an iterative think-act-rethink chain that treats medical images as interactive objects, with visual attention analysis showing that reasoning rounds increasingly anchor to clinically critical regions. DCI~\cite{xu2026dual} addresses confounding biases through a causal framework that disentangles true causal effects from spurious cross-modal shortcuts. Zafar et~al.~\cite{zafar2026beyond} propose counterfactual evaluation using real, blank, and shuffled images, introducing metrics such as Visual Reliance Score (VRS), Image Sensitivity (IS), and Hallucinated Visual Reasoning Rate (HVRR) to measure genuine visual dependence. HALT-MedVQA~\cite{wu2024hallucination} constructs hallucination stress tests through fake questions, ``None of the Above'' choices, and mismatched image substitutions. Additional factors affecting medical VLM reliability include resolution sensitivity~\cite{chen2025impact} and multi-image reasoning demands~\cite{yu2025medframeqa}. These findings motivate our visual grounding analysis as a necessary complement to accuracy reporting.

\section{Methodology}
\label{sec:methodology}

\subsection{Pipeline Overview}

\begin{figure}
\centering
\includegraphics[width=\textwidth]{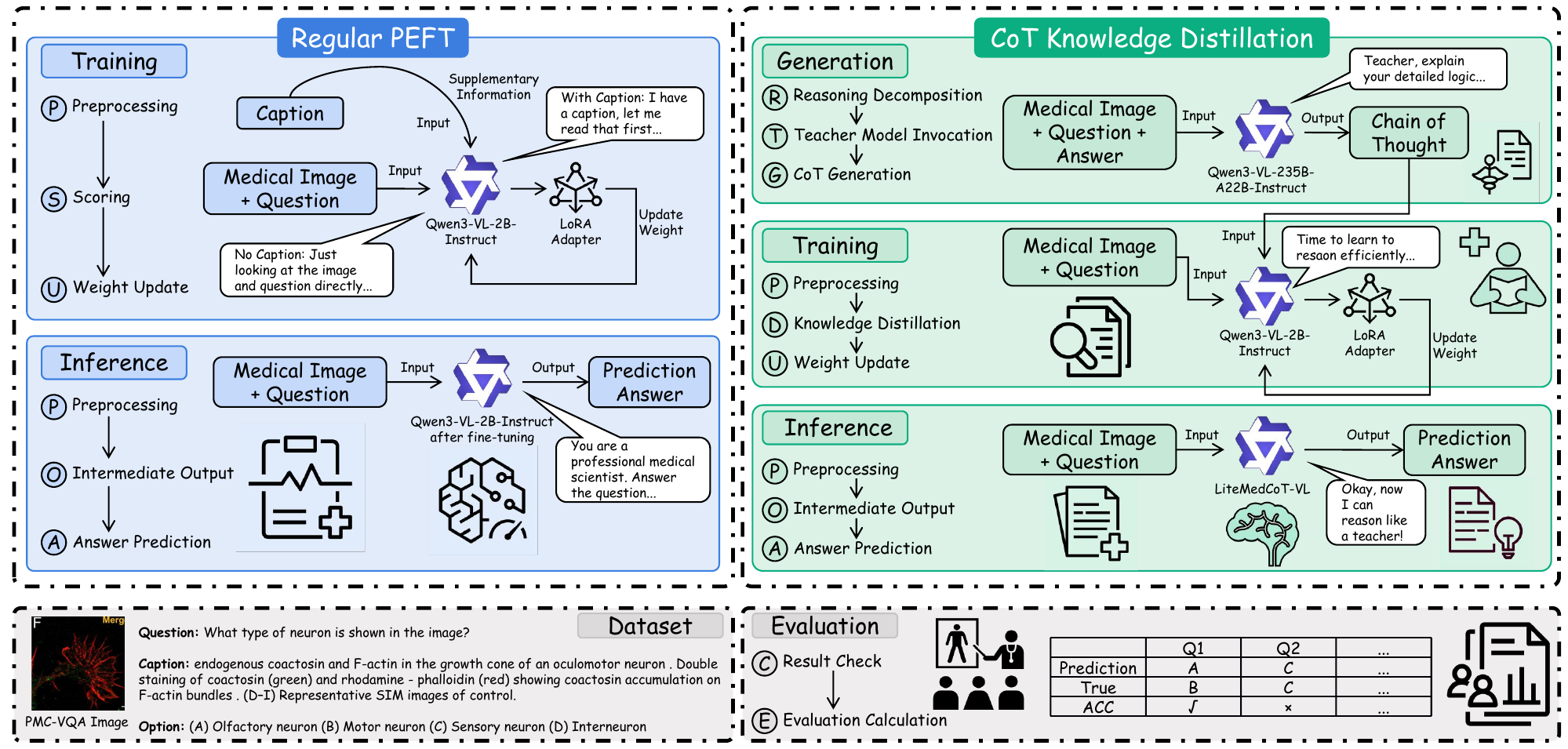}
\caption{Overview of the LiteMedCoT-VL pipeline. Training samples are processed through prompt engineering, chain-of-thought explanation generation from a large teacher model, and LoRA-based fine-tuning of compact student models.}
\label{fig:pipeline}
\end{figure}

LiteMedCoT-VL adapts compact vision-language models for medical visual question answering through chain-of-thought knowledge distillation and parameter-efficient fine-tuning. Figure~\ref{fig:pipeline} provides an overview of the pipeline. It consists of three stages. First, a large teacher model generates chain-of-thought explanations for training samples. Second, these explanations are injected into the training data. Third, the compact student model is fine-tuned using LoRA on the enriched dataset. Model adaptation applies LoRA to the query, key, value, and output projections (\texttt{q\_proj}, \texttt{k\_proj}, \texttt{v\_proj}, \texttt{o\_proj}) with rank $r=32$, scaling factor $\alpha=64$, and dropout rate 0.05. These LoRA modules are inserted into every attention layer of the language model backbone, while the vision encoder remains frozen. Freezing the vision encoder preserves the pretrained visual representations and reduces the number of trainable parameters, which is critical under the memory constraints of portable deployment. The full pipeline uses standardized prompts and deterministic evaluation.

\subsection{Prompt Design and Inference}

The pipeline uses two core prompt families for answer-only inference and a third for chain-of-thought training. All prompts enforce strict output formatting to eliminate parsing ambiguity; the exact prompt texts are provided in Appendix~\ref{sec:appendix-prompts}.

Our primary evaluation protocol provides the model with only the image and question, requiring it to output a single uppercase letter without explanation. We adopt this no-caption setting as the default because it simulates the clinical workflow in which a physician interprets a medical image directly from visual evidence, without an accompanying radiology report. This scenario is common in primary care, emergency departments, and point-of-care settings where a formal written report is not immediately available.

The no-caption setting also provides a more rigorous evaluation. Caption text may contain answer-relevant cues that shortcut visual reasoning, inflating accuracy without genuine image understanding. In PMC-VQA specifically, captions can correlate with answer labels, creating a risk of data leakage. By defaulting to no-caption inference, we ensure that reported performance reflects visual reasoning rather than textual priors. Caption-aware results serve as an upper-bound reference and are reported in the ablation study (Section~\ref{sec:results}).

We employ next-token probability scoring rather than autoregressive generation to eliminate decoding stochasticity. For each sample, the model processes the image and prompt, and we extract logits at the final token position for the four candidate labels A, B, C, and D, including both bare and space-prefixed tokenizations. The label with the highest logit is selected as the prediction. This approach ensures deterministic, reproducible evaluation.

\section{Experiments}
\label{sec:experiments}

\subsection{Datasets}

The PMC-VQA dataset~\cite{zhang2023pmc} serves as the primary evaluation benchmark. The dataset provides a training split of 152,603 samples and a test split of 2,000 samples. All fine-tuning is performed exclusively on the training split; all reported accuracy metrics are computed on the held-out test split. Detailed dataset statistics are provided in Table~\ref{tab:dataset} in Appendix~\ref{sec:appendix-tables}.

\begin{figure}
\centering
\includegraphics[width=0.75\textwidth]{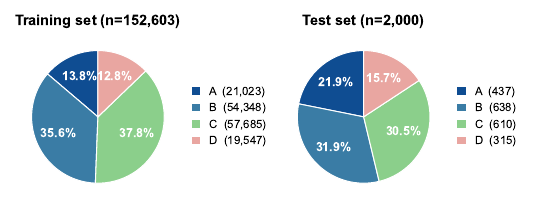}
\caption{Answer label distribution in the PMC-VQA training and test sets. Options B and C dominate both splits, accounting for 73.5\% of training and 62.4\% of test samples.}
\label{fig:dataset_dist}
\end{figure}

The answer distribution across options A--D is imbalanced in both splits, as shown in Figure~\ref{fig:dataset_dist}. In the training set, options B and C together account for 73.5\% of samples, with 35.6\% and 37.8\% for B and C respectively, while options A and D each represent less than 14\%. The test set exhibits a similar but less pronounced skew, with B and C comprising 62.4\% of samples, 31.9\% and 30.5\% for each, and A and D accounting for 21.9\% and 15.8\% respectively. This imbalance introduces an answer-position bias during fine-tuning, as the model learns to favor the more frequent answer positions.

For error analysis, we define nine question type categories based on the semantic content of the questions: \emph{modality} for imaging technique identification, \emph{anatomy} for anatomical structure identification, \emph{color/label} for visual marker interpretation, \emph{diagnosis} for disease or condition identification, \emph{counting} for enumeration tasks, \emph{comparison} for relational reasoning, \emph{temporal} for staging or progression, \emph{procedure} for treatment or intervention, and \emph{other} for questions matching no category keywords. These categories are not provided by the PMC-VQA dataset; we define them heuristically using keyword matching on the question text to enable fine-grained error analysis. A question may match multiple categories. The complete keyword definitions are provided in Table~\ref{tab:question_types} in Appendix~\ref{sec:appendix-categories}.

\subsection{Baselines}

We select baselines to cover both the published PMC-VQA leaderboard and representative compact open-source models. The published methods include PMC-CLIP~\cite{radford2021learning}, BLIP-2~\cite{li2023blip}, Open-Flamingo~\cite{awadalla2023openflamingo}, LLaVA-Med~\cite{li2023llava}, and MedICap-GPT-4~\cite{zhang2023pmc}, which are chosen because they represent the best-reported results on the PMC-VQA benchmark. A trained variant, MedICaP-PMCVQA-GPT-4, is also included as the strongest supervised baseline. Additionally, we evaluate three compact open-source vision-language models under zero-shot inference to establish cross-architecture baselines: InternVL2-2B~\cite{chen2024far}, SmolVLM2-2.2B~\cite{marafioti2025smolvlm}, and Phi-3.5-vision~\cite{abdin2024phi3}, selected as current mainstream compact models suitable for portable device deployment. For our fine-tuning experiments, we use Qwen3-VL-2B-Instruct~\cite{bai2025qwen3} and Qwen3-VL-4B-Instruct~\cite{bai2025qwen3} as student models, with Qwen3-VL-235B-A22B-Instruct~\cite{bai2025qwen3} as the teacher for chain-of-thought generation.

\subsection{Implementation}

Hardware and software specifications are provided in Appendix~\ref{sec:appendix-setup}. Fine-tuning applies LoRA with rank $r=32$, scaling factor $\alpha=64$, and dropout 0.05 to the query, key, value, and output projections. The training data is split into 51 chunks of approximately 3{,}000 samples each, processed sequentially with adapter weights carried forward. Each chunk is trained for 1 epoch, yielding an effective total of 1 epoch over the full dataset. Training uses the AdamW optimizer with a learning rate of $2\times10^{-4}$ and 100 warmup steps under a linear schedule. The effective batch size is 8, achieved via per-device batch size 1 and gradient accumulation over 4 steps. Gradient norms are clipped to 1.0. All images are processed at their native resolution.

For chain-of-thought distillation, the teacher model Qwen3-VL-235B-A22B-Instruct generates explanations via API for 152,601 of 152,603 training samples, achieving 99.99\% coverage. The average explanation length is 147 words with a median of 139 and a range of 5--438. Representative teacher explanations are provided in Appendix~\ref{sec:appendix-cot}.

Inference uses deterministic next-token logit scoring in bfloat16, with no sampling. To prevent test contamination, the teacher generates explanations exclusively for training split samples. This work uses publicly available benchmark data and does not involve new human-subject data collection.

\section{Results}
\label{sec:results}

\subsection{Comparison Results}

\begin{figure}
\centering
\includegraphics[width=\textwidth]{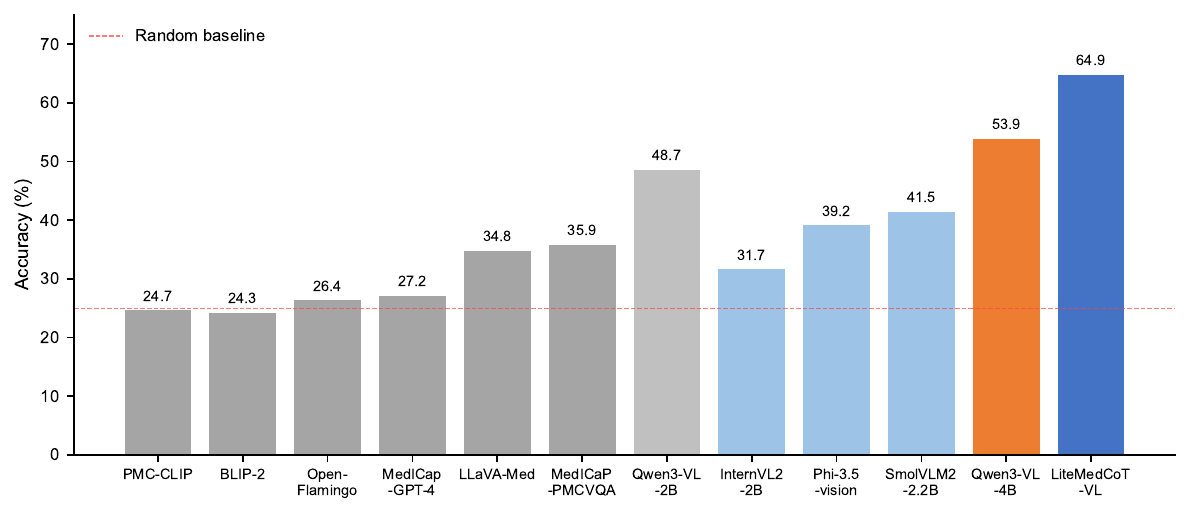}
\caption{Accuracy of evaluated models on the PMC-VQA test set. The horizontal dashed line indicates the 25\% random baseline for four-option multiple choice. Our LiteMedCoT-VL achieves 64.9\%, outperforming all baselines.}
\label{fig:comparison}
\end{figure}

Figure~\ref{fig:comparison} reports accuracy for all evaluated models on the PMC-VQA test set; the full numerical data are provided in Table~\ref{tab:comparison_app} in Appendix~\ref{sec:appendix-comparison}. LiteMedCoT-VL achieves 64.9\% accuracy, outperforming all published baselines. The base Qwen3-VL-2B achieves 48.7\% without fine-tuning. The fine-tuned 2B model exceeds the zero-shot Qwen3-VL-4B at 53.9\% by 11.0 percentage points. The zero-shot SmolVLM2-2.2B achieves 41.5\%, already surpassing all published baselines. InternVL2-2B scores 31.7\% zero-shot.

The published baseline numbers, from PMC-CLIP through MedICaP-PMCVQA-GPT-4, are taken from~\cite{zhang2023pmc} and were obtained under different prompt designs and inference protocols. The compact open-source models are evaluated under our standardized deterministic PPL scoring protocol. Bootstrap 95\% confidence intervals with 10,000 resamples for our evaluated models are: Qwen3-VL-2B $48.7\%$ $[46.6, 50.9]$, Qwen3-VL-4B $53.8\%$ $[51.6, 56.0]$, InternVL2-2B $31.7\%$ $[29.6, 33.8]$, SmolVLM2-2.2B $41.4\%$ $[39.3, 43.5]$, Phi-3.5-vision $39.1\%$ $[37.0, 41.3]$. Head-to-head comparison of all methods under matched inference protocols remains desirable but is constrained by the availability of model weights and computational resources.

\subsection{Ablation Study}

Table~\ref{tab:ablation} presents the ablation results comparing different training configurations. The base Qwen3-VL-2B achieves 48.7\% with default settings. Fine-tuning with answer-only supervision and no caption improves performance to 54.2\%. Adding caption-aware training further increases accuracy to 60.5\%. The full LiteMedCoT-VL pipeline, which injects chain-of-thought reasoning from Qwen3-VL-235B-A22B-Instruct into the training data, achieves 64.9\%, a 4.4 percentage point improvement over caption-aware training alone. Each configuration is evaluated deterministically on the same 2,000-sample test set with fixed inference settings.

\begin{table}
\centering
\caption{Ablation of training configurations on Qwen3-VL-2B. Starting from the default 48.7\%, no-caption fine-tuning, caption-aware training, and chain-of-thought distillation are applied incrementally. Each stage adds a supervision signal, with reasoning distillation providing the largest gain.}
\label{tab:ablation}
\begin{tabular}{lcc}
\toprule
Configuration & Accuracy  & $\Delta$  \\
\midrule
Qwen3-VL-2B              & 48.7 & -- \\
\quad + no-caption fine-tuning   & 54.2 & +5.5 \\
\quad + caption-aware training   & 60.5 & +6.3 \\
\quad + CoT distillation (LiteMedCoT-VL) & \textbf{64.9} & +4.4 \\
\bottomrule
\end{tabular}
\end{table}

Fine-tuning introduces an answer-position bias: the fine-tuned model achieves higher accuracy on options B at 61.0\% and C at 71.6\%, compared to A at 36.6\% and D at 31.1\%. This pattern reflects the imbalanced answer distribution in the PMC-VQA training set, where B and C account for 73.5\% of samples. The bias is less pronounced in the default Qwen3-VL-4B, which ranges from 45.1\% to 58.9\% across positions.

\subsection{Visual Grounding Analysis}

\begin{figure}
\centering
\includegraphics[width=0.85\textwidth]{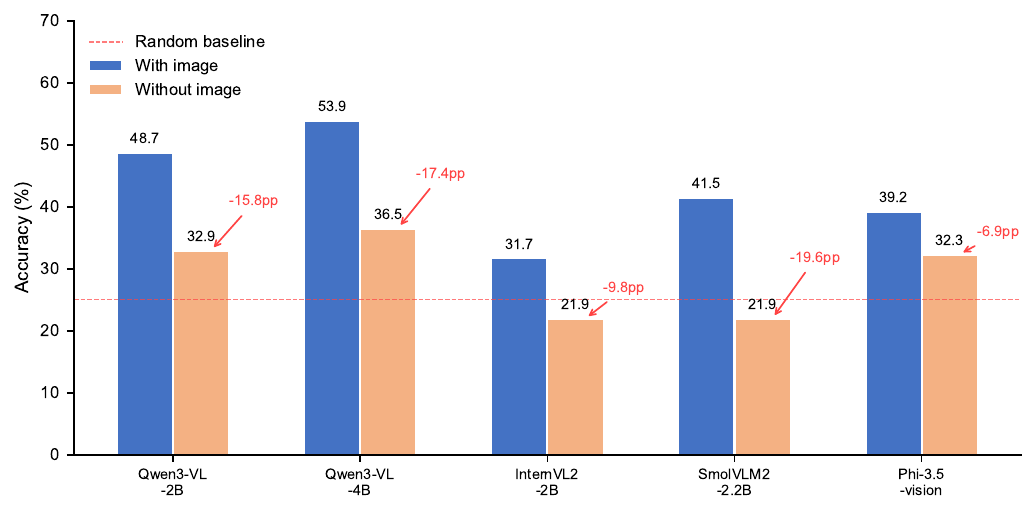}
\caption{Image ablation results across all baseline models. Removing images causes substantial accuracy drops for all models, confirming genuine visual reliance. Fine-tuned variants are excluded because fine-tuning may alter the model's reliance on visual versus textual cues.}
\label{fig:ablation_visual}
\end{figure}

To assess whether the models genuinely rely on visual information, we conduct an image ablation study across all baseline models. Figure~\ref{fig:ablation_visual} presents the results of running inference with images removed, using only the question and options text. All models show substantial performance degradation when images are withheld. Qwen3-VL-2B drops from 48.7\% to 32.9\%, a 15.8pp decrease, and Qwen3-VL-4B drops from 53.9\% to 36.5\%, a 17.4pp decrease, confirming that both models leverage visual information for prediction. SmolVLM2-2.2B shows the largest drop of 19.6pp, while Phi-3.5-vision shows a more modest decrease of 6.9pp. The no-image accuracies for InternVL2-2B and SmolVLM2-2.2B fall to 21.9\%, below the 25\% random baseline, indicating that these models default to predicting the most frequent answer position when images are removed rather than exploiting textual cues. In contrast, Qwen3-VL-2B, Qwen3-VL-4B, and Phi-3.5-vision maintain no-image accuracies above the random baseline at 32.9\%, 36.5\%, and 32.3\% respectively, suggesting some exploitation of textual cues within questions and answer options, consistent with findings from MIRAGE~\cite{asadi2026mirage}. The exact numerical data are provided in Table~\ref{tab:ablation_visual} in Appendix~\ref{sec:appendix-ablation}.

The fine-tuned variants, namely LiteMedCoT-VL, no-caption FT, and caption FT, are excluded from this ablation because the fine-tuning procedure itself may alter the model's reliance on visual versus textual cues. Specifically, fine-tuning on answer-only supervision with imbalanced answer distributions can amplify position bias, confounding the interpretation of the no-image experiment.

\subsection{Error Analysis by Question Type}

To understand where chain-of-thought distillation yields the greatest gains, we categorize test questions by type using keyword matching; definitions are provided in Appendix~\ref{sec:appendix-categories}. Figure~\ref{fig:error_analysis} reports results for all evaluated models; complete numerical data appear in Table~\ref{tab:error_analysis_app} in Appendix~\ref{sec:appendix-error}. Among the zero-shot baselines, Qwen3-VL-4B consistently outperforms the smaller models, while InternVL2-2B shows the weakest performance across most categories. Fine-tuning yields the largest gains on \emph{anatomy} at +24.7pp and \emph{color/label} at +19.5pp, both categories that require visual interpretation of imaging features. The \emph{procedure} category achieves the highest accuracy across all configurations, ranging from 35.8\% to 82.1\%. The \emph{comparison} category remains the most challenging for most models, with InternVL2-2B and SmolVLM2-2.2B scoring below 30\%. The fine-tuned model improves accuracy across all answer positions, with position A showing the largest gain of 22.7pp and positions B, C, and D each improving by 13--16pp.

\begin{figure}
\centering
\includegraphics[width=\textwidth]{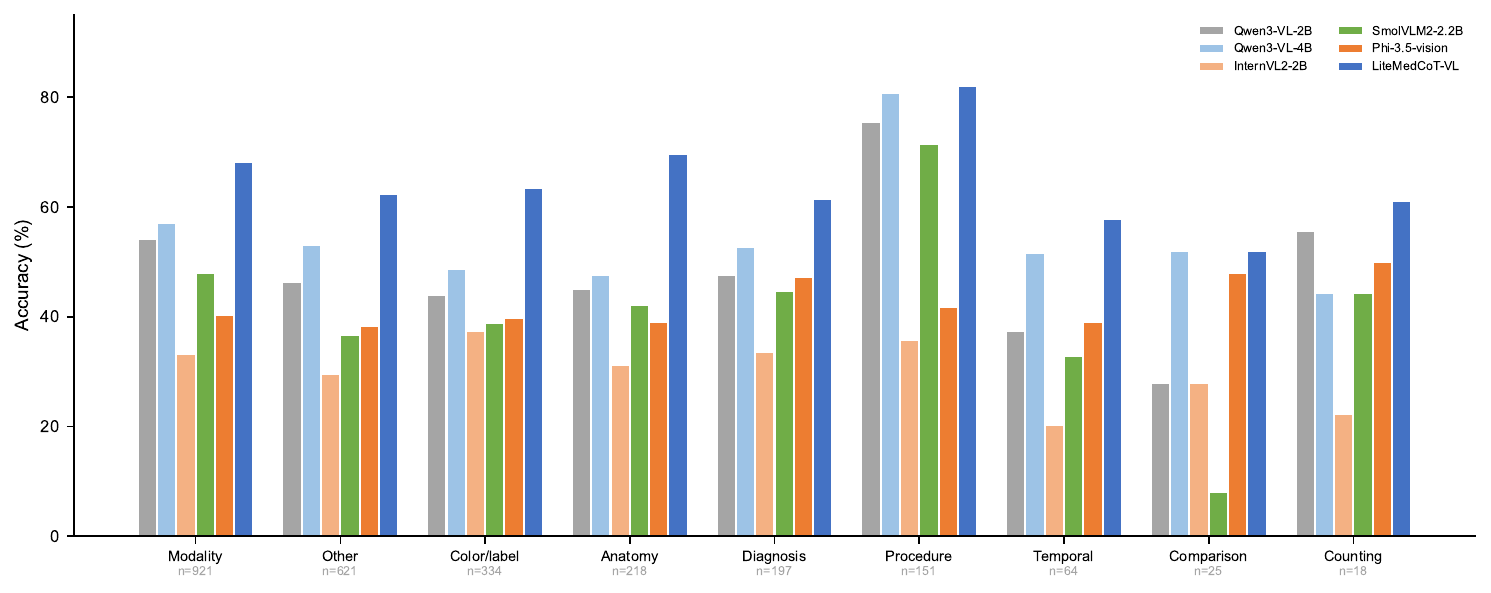}
\caption{Per-category accuracy on the PMC-VQA test set. Sample sizes per category are shown below the x-axis. LiteMedCoT-VL achieves the largest gains on anatomy and color/label relative to the 2B default.}
\label{fig:error_analysis}
\end{figure}

\section{Discussion}
\label{sec:discussion}

The results demonstrate that chain-of-thought distillation from a 235B teacher improves a 2B student model by 16.2 percentage points on PMC-VQA, with 4.4pp of that gain attributable specifically to CoT distillation beyond caption-aware training. This finding aligns with the broader trend of reasoning transfer in medical VQA, though the mechanisms differ across approaches. Step-CoT~\cite{fan2026step} leverages expert-curated structured annotations with graph-based focusing, achieving gains through annotation quality. CheXthought~\cite{sharma2026chexthought} shows that radiologist-authored traces with visual attention hints surpass VLM-generated CoT in factual accuracy. MedE2~\cite{huang2025elicit} combines demonstration elicitation with Direct Preference Optimization. Trajectory-aware GRPO~\cite{gulluk2026grpo} applies process rewards based on reasoning-step similarity, and CARE~\cite{du2026care} decomposes VQA into specialized sub-modules trained with reinforcement learning. LiteMedCoT-VL offers a simpler, single-stage recipe that does not require expert annotations or multi-stage pipelines, though it does not match the structured grounding of agentic frameworks like CARE. The trade-off between recipe simplicity and reasoning fidelity warrants further investigation.

The 11.0 percentage point gap between LiteMedCoT-VL at 64.9\% and the zero-shot Qwen3-VL-4B at 53.9\% indicates that reasoning capability, rather than parameter count alone, drives performance on this benchmark. The caption-aware results further illuminate this finding: the 6.3pp improvement from caption inclusion, from 54.2\% to 60.5\%, suggests that textual context complements visual features, analogous to how a radiology report supplements a clinician's direct image interpretation. However, this gain may partly reflect shortcut learning from caption--answer correlations in the training data rather than genuine multimodal integration.

The answer-position bias observed after fine-tuning, with accuracy ranging from 31.1\% for option D to 71.6\% for option C, reflects the imbalanced answer distribution in the PMC-VQA training set. This bias is less pronounced in the default Qwen3-VL-4B at 45.1\%--58.9\%, suggesting that fine-tuning amplifies sensitivity to label frequency. Mitigation strategies such as balanced sampling, label smoothing, or focal loss merit exploration in future iterations.

\section{Limitations}
\label{sec:limitations}

Evaluation is limited to PMC-VQA and the Qwen3-VL model family. Cross-dataset generalization to SLAKE~\cite{liu2021slake}, PathVQA~\cite{he2020pathvqa}, and VQA-RAD~\cite{lau2018dataset}, and cross-architecture distillation remain open. The single-image evaluation does not capture multi-image clinical reasoning~\cite{yu2025medframeqa} or resolution sensitivity~\cite{chen2025impact}.

The fine-tuned variants are excluded from the no-image ablation because fine-tuning may alter the model's reliance on visual versus textual cues. More rigorous grounding verification of the distilled model remains necessary. The counterfactual evaluation framework of Zafar et~al.~\cite{zafar2026beyond}, which introduces VRS, IS, and HVRR metrics using real, blank, and shuffled images, provides a principled approach. HALT-MedVQA~\cite{wu2024hallucination} stress tests using fake questions, ``None of the Above'' choices, and image swaps offer complementary diagnostic signals. The entropy-guided regrounding of MedVR~\cite{jiang2026medvr}, the iterative think-act-rethink chains of ViTAR~\cite{chen2025think}, and the causal deconfounding framework of DCI~\cite{xu2026dual} provide additional avenues for grounding verification.

All results are reported from single training runs. The deterministic inference protocol eliminates decoding variance, but training variance from data ordering and random initialization is not quantified. The published baseline numbers are taken from prior work under different inference protocols, limiting the fairness of direct comparison.

We have not measured latency, memory footprint, throughput, or energy consumption on representative portable hardware. Quantization effects on accuracy are also unexplored. These measurements are necessary to substantiate practical deployment claims.

All experiments use curated benchmark data under offline evaluation. No clinical deployment validation or expert review has been conducted. Teacher-generated explanations are assessed only by coverage and length statistics, not by clinical expert evaluation. The faithfulness of distilled reasoning chains to the teacher's actual visual attention remains an open question; CheXthought~\cite{sharma2026chexthought} demonstrates that expert-verified traces with visual attention annotations yield more grounded reasoning, suggesting that lightweight clinician audits of teacher outputs could improve distillation quality.

\section{Conclusion}
\label{sec:conclusion}

This work demonstrates that chain-of-thought knowledge distillation from a 235B teacher can improve the performance of compact vision-language models on medical visual question answering. LiteMedCoT-VL achieves 64.9\% on PMC-VQA, exceeding the 4B model at 53.9\% and all published baselines. The distillation from Qwen3-VL-235B-A22B-Instruct improves student accuracy from 48.7\% to 64.9\%, indicating that reasoning capability transfers effectively across model scales. A 2B model with LoRA adaptation can outperform a model with twice the parameters, reducing the compute and memory requirements for resource-constrained clinical settings. Visual grounding verification, multi-dataset evaluation, efficiency measurements on portable hardware, and clinical validation remain necessary to establish the reliability of these findings for deployment.

\newpage

\appendix
\section{Appendix}

\subsection{System Prompts}
\label{sec:appendix-prompts}

The complete prompts used in the pipeline are reproduced below, corresponding to the three prompt families described in Section~\ref{sec:methodology}.

\subsubsection*{No-Caption Inference Prompt}

\textbf{System prompt}
\begin{verbatim}
You are a professional medical scientist. Answer the choice
question based strictly on the image.
STRICT OUTPUT FORMAT:
1. You MUST output ONLY a single uppercase letter: A, B, C, or D.
2. DO NOT output the full option text.
3. DO NOT output phrases like 'The answer is'.
4. NO explanation, NO reasoning, NO punctuation.
Example Output:
A
\end{verbatim}

\textbf{User message format}
\begin{verbatim}
Question: [question text]
Options:
A. [option A]
B. [option B]
C. [option C]
D. [option D]
\end{verbatim}

\subsubsection*{Caption-Aware Inference Prompt}

\textbf{System prompt}
\begin{verbatim}
You are a professional medical scientist. Answer the choice
question based strictly on the image and the caption.
STRICT OUTPUT FORMAT:
1. You MUST output ONLY a single uppercase letter: A, B, C, or D.
2. DO NOT output the full option text.
3. DO NOT output phrases like 'The answer is'.
4. NO explanation, NO reasoning, NO punctuation.
Example Output:
A
\end{verbatim}

\textbf{User message format}
\begin{verbatim}
Image caption: [caption text]
Question: [question text]
Options:
A. [option A]
B. [option B]
C. [option C]
D. [option D]
\end{verbatim}

\subsubsection*{Chain-of-Thought Training Prompt}

This prompt is used during SFT training with distilled explanations from the teacher model (Section~\ref{sec:methodology}).

\textbf{System prompt}
\begin{verbatim}
You are a professional medical scientist. Answer the choice
question based strictly on the image.
OUTPUT FORMAT:
1. First, provide your reasoning and analysis based on the image.
2. Then output on a new line exactly: Answer: <LETTER>.
3. The letter MUST be A, B, C, or D.
Example Output:
Explanation: [Your detailed analysis of the image findings]
Answer: A
\end{verbatim}

\subsection{Hardware and Software Setup}
\label{sec:appendix-setup}

Table~\ref{tab:hardware} specifies the hardware and software environment used for all experiments (Section~\ref{sec:experiments}).

\begin{table}[H]
\centering
\caption{Hardware and software specifications.}
\label{tab:hardware}
\begin{tabular}{ll}
\toprule
Component & Specification \\
\midrule
GPU & NVIDIA GeForce RTX 5090, 32\,GB VRAM \\
CPU & 25 vCPU Intel Xeon Platinum 8470Q \\
RAM & 120\,GB \\
OS & Ubuntu 22.04 \\
Python & 3.12 \\
PyTorch & 2.8.0 \\
CUDA & 12.8 \\
Transformers & 4.45+ \\
Attention & Flash Attention 2 \\
\bottomrule
\end{tabular}
\end{table}

\subsection{Teacher Chain-of-Thought Examples}
\label{sec:appendix-cot}

The following examples illustrate the quality and diversity of chain-of-thought explanations generated by the teacher model Qwen3-VL-235B-A22B-Instruct for PMC-VQA training samples (Section~\ref{sec:methodology}).

\subsubsection*{Example 1: Color identification}

\textbf{Question:} What color is used to label the Golgi complexes in the image?\\
\textbf{Answer:} B (Red)

\textbf{Teacher explanation:} The image depicts a three-dimensional reconstruction of a eukaryotic cell, with various organelles labeled using distinct colors. The nucleus is clearly visible in the center, rendered in light blue, which is a common color choice for nuclear labeling in such visualizations. Surrounding the nucleus, the green structures represent the endoplasmic reticulum (ER), identifiable by their extensive network and tubular morphology. The Golgi complex, located near the nucleus and typically positioned adjacent to the ER, is shown as a cluster of small, stacked, red-colored vesicles. This red labeling corresponds to the Golgi apparatus, which is often highlighted in red in fluorescent microscopy or 3D reconstructions to distinguish it from other organelles. Therefore, based on the visual features---specifically the red-stained vesicular structure adjacent to the nucleus and ER---the correct answer is B: Red.

\subsubsection*{Example 2: Modality identification}

\textbf{Question:} What type of imaging was used to demonstrate acute infarct?\\
\textbf{Answer:} C (Diffusion weighted image)

\textbf{Teacher explanation:} The image shows two axial brain slices with hyperintense (bright) lesions in the cerebral cortex and subcortical regions, particularly prominent in the right hemisphere. These bright signals are characteristic of restricted diffusion, which is a hallmark of acute ischemic stroke. Diffusion-weighted imaging (DWI) is highly sensitive to cytotoxic edema that occurs within minutes of ischemia, making it the most effective MRI sequence for detecting acute infarcts. The presence of these bright lesions on DWI, coupled with the typical location and acute clinical context implied by the question, confirms that the imaging modality used was diffusion-weighted imaging. Time-of-flight MR (A) is used for vascular imaging, not tissue infarct detection; DSA (B) is an invasive angiographic technique primarily for evaluating vessel patency, not parenchymal changes. Therefore, the correct answer is C: Diffusion weighted image.

\subsubsection*{Example 3: Anatomical structure identification}

\textbf{Question:} What is shown in the image?\\
\textbf{Answer:} C (Transverse section anatomy of the storage root)

\textbf{Teacher explanation:} The image displays a transverse (cross) section of a plant structure, characterized by a central vascular bundle surrounded by parenchyma cells, which is typical of a storage root. The presence of a large central stele with phloem and xylem arranged in a radial pattern, along with thick-walled storage parenchyma cells, indicates a modified root adapted for storage. This anatomy is consistent with a transverse section of a storage root, such as those found in carrots or beets. The two upward-pointing arrows likely highlight the vascular bundles, further supporting the identification of a root structure. The overall organization lacks the epidermis and stomata seen in leaves, and the arrangement differs from stem anatomy, which typically shows vascular bundles in a ring or scattered pattern. Therefore, the correct answer is C: Transverse section anatomy of the storage root.

\subsection{Dataset Statistics}
\label{sec:appendix-tables}

Table~\ref{tab:dataset} reports the PMC-VQA dataset statistics referenced in Section~\ref{sec:experiments}.

\begin{table}[H]
\centering
\caption{PMC-VQA dataset statistics by answer label.}
\label{tab:dataset}
\begin{tabular}{lrr}
\toprule
 & Training & Test \\
\midrule
Total samples & 152,603 & 2,000 \\
Answer A & 21,023 & 437 \\
Answer B & 54,348 & 638 \\
Answer C & 57,685 & 610 \\
Answer D & 19,547 & 315 \\
\bottomrule
\end{tabular}
\end{table}

\subsection{Question Type Definitions}
\label{sec:appendix-categories}

Table~\ref{tab:question_types} defines the nine question categories used in the error analysis (Section~\ref{sec:results}). Categories are assigned by keyword matching on the question text; a question may match multiple categories.

\begin{table}[H]
\centering
\caption{Question type categories and their keyword definitions.}
\label{tab:question_types}
\small
\begin{tabular}{lp{9cm}}
\toprule
Category & Keywords \\
\midrule
Modality & imaging, modality, technique, scan, MRI, CT, X-ray, radiograph, ultrasound, microscope, photograph, fluorescence, PET, SPECT, endoscope, histology, pathology \\
Anatomy & organ, structure, anatomy, region, location, lobe, artery, vein, nerve, bone, muscle, tissue, cell, membrane, cortex, nucleus, ventricle \\
Color/label & color, label, labelled, labeled, arrow, highlight, indicate, mark, point \\
Diagnosis & diagnosis, disease, condition, pathological, abnormal, finding, appearance, suggest, consistent with, likely \\
Counting & how many, number of, count, quantity, multiple \\
Comparison & compare, difference, differ, similar, versus, vs, than, more, less, larger, smaller \\
Temporal & stage, phase, progress, develop, time, course, acute, chronic, early, late, before, after \\
Procedure & procedure, treatment, surgery, intervention, therapy, approach, technique, method \\
Other & no keyword match \\
\bottomrule
\end{tabular}
\end{table}

\subsection{Comparison Results}
\label{sec:appendix-comparison}

Table~\ref{tab:comparison_app} provides the exact accuracy values for all evaluated models corresponding to Figure~\ref{fig:comparison} (Section~\ref{sec:results}).

\begin{table}[H]
\centering
\caption{Accuracy of all evaluated models on the PMC-VQA test set.}
\label{tab:comparison_app}
\begin{tabular}{lc}
\toprule
Model & Accuracy (\%) \\
\midrule
PMC-CLIP                           & 24.7  \\
BLIP-2                             & 24.3  \\
Open-Flamingo                      & 26.4  \\
MedICap-GPT-4                      & 27.2  \\
LLaVA-Med                          & 34.8  \\
MedICaP-PMCVQA-GPT-4               & 35.9  \\
Qwen3-VL-2B                        & 48.7  \\
InternVL2-2B                       & 31.7  \\
Phi-3.5-vision                     & 39.2  \\
SmolVLM2-2.2B                      & 41.5  \\
Qwen3-VL-4B                        & 53.9  \\
\textbf{LiteMedCoT-VL}             & \textbf{64.9} \\
\bottomrule
\end{tabular}
\end{table}

\subsection{Ablation and Visual Grounding}
\label{sec:appendix-ablation}

Table~\ref{tab:ablation_visual} provides the image ablation accuracy corresponding to Figure~\ref{fig:ablation_visual} (Section~\ref{sec:results}).

\begin{table}[H]
\centering
\caption{Image ablation accuracy with and without images for all baseline models.}
\label{tab:ablation_visual}
\begin{tabular}{lccc}
\toprule
Model & With image (\%) & Without image (\%) & $\Delta$ (pp) \\
\midrule
Qwen3-VL-2B     & 48.7  & 32.9  & $-$15.8  \\
Qwen3-VL-4B     & 53.9  & 36.5  & $-$17.4  \\
InternVL2-2B    & 31.7  & 21.9  & $-$9.8   \\
SmolVLM2-2.2B   & 41.5  & 21.9  & $-$19.6  \\
Phi-3.5-vision  & 39.2  & 32.3  & $-$6.9   \\
\bottomrule
\end{tabular}
\end{table}

\subsection{Error Analysis Data}
\label{sec:appendix-error}

Table~\ref{tab:error_analysis_app} provides per-category accuracy values corresponding to Figure~\ref{fig:error_analysis} in the main text.

\begin{table}[H]
\centering
\caption{Per-category accuracy on the PMC-VQA test set.}
\label{tab:error_analysis_app}
\resizebox{\textwidth}{!}{\begin{tabular}{lccccccccc}
\toprule
Model & Modality & Other & Color/label & Anatomy & Diagnosis & Procedure & Temporal & Comparison & Counting \\
 & $n$=921 & $n$=621 & $n$=334 & $n$=218 & $n$=197 & $n$=151 & $n$=64 & $n$=25 & $n$=18 \\
\midrule
Qwen3-VL-2B           & 54.2 & 46.4 & 44.0 & 45.0 & 47.7 & 75.5 & 37.5 & 28.0 & 55.6 \\
Qwen3-VL-2B-NoCaption & 60.6 & 52.0 & 47.0 & 53.7 & 52.8 & 79.5 & 42.2 & 28.0 & 44.4 \\
Qwen3-VL-2B-Caption   & 63.8 & 59.1 & 53.0 & 57.3 & 62.4 & 73.5 & 53.1 & 48.0 & 66.7 \\
Qwen3-VL-4B           & 57.1 & 53.0 & 48.8 & 47.7 & 52.8 & 80.8 & 51.6 & 52.0 & 44.4 \\
InternVL2-2B          & 33.3 & 29.5 & 37.4 & 31.2 & 33.5 & 35.8 & 20.3 & 28.0 & 22.2 \\
SmolVLM2-2.2B         & 47.9 & 36.6 & 38.9 & 42.2 & 44.7 & 71.5 & 32.8 & 8.0  & 44.4 \\
Phi-3.5-vision        & 40.3 & 38.3 & 39.8 & 39.0 & 47.2 & 41.7 & 39.1 & 48.0 & 50.0 \\
\textbf{LiteMedCoT-VL}         & \textbf{68.3} & \textbf{62.3} & \textbf{63.5} & \textbf{69.7} & 61.4 & \textbf{82.1} & \textbf{57.8} & \textbf{52.0} & 61.1 \\
\bottomrule
\end{tabular}}
\end{table}

\clearpage
\bibliographystyle{unsrtnat}
\bibliography{references}

\end{document}